\documentclass[12pt]{article}

\usepackage{sbc-template}
\usepackage{graphicx,url}
\usepackage[utf8]{inputenc}
\usepackage[T1]{fontenc}
\usepackage[brazil,english]{babel}
\usepackage{multirow}
\usepackage{subfig}
\usepackage{tcolorbox}
\usepackage{listings}
\usepackage{url}
\usepackage{hyperref}

\title{Integrating Large Language Models and Graph Convolutional Networks for Semi-Supervised Image Classification}

\author{
    Camila Piscioneri Magalhães\inst{1},
    Lucas Pascotti Valem\inst{1}
}

\address{
Institute of Mathematics and Computer Science (ICMC) \\
University of São Paulo (USP) \\
São Carlos -- SP -- Brazil
\email{piscioneri@usp.br, lucas@icmc.usp.br}
}

\begin{document}

\maketitle

\begin{center}
\vspace{-2mm}
\fbox{\parbox{0.92\columnwidth}{\centering\small
\textbf{Preprint.} Accepted at the Workshop de Trabalhos de Alunos de Gradua\c{c}\~ao (WTAG) of the Brazilian Symposium on Databases (SBBD~2026).}}
\vspace{2mm}
\end{center}

\begin{abstract}
While the growing availability of image data has driven significant advances, labeling datasets remains costly and time-consuming. Therefore, semi-supervised approaches such as Graph Convolutional Networks (GCNs), which learn from both labeled and unlabeled data, have emerged as a promising solution. One of the primary challenges in applying GCNs to image classification is graph construction, since, unlike in citation networks or similar domains, images typically do not come with a predefined structural representation. For visual data, most studies construct graphs based on the similarity between feature vectors from pretrained deep learning backbones, typically by employing kNN or reciprocal kNN algorithms. Although Large Language Models (LLMs) have shown remarkable capability in capturing high-level semantics, their integration with GCNs for image classification remains underexplored. Aiming to fill this gap, our approach uses a Vision Language Model (VLM) to generate textual image descriptions, which are then processed by an LLM to estimate semantic similarity scores between connected images. These scores guide the pruning of edges in kNN and reciprocal kNN graphs, filtering out semantically irrelevant neighbors.
Experimental results reveal that leveraging LLMs for graph refinement can improve classification accuracy, particularly for kNN graphs and some backbones. The source code is publicly available at \href{http://gcnllm.lucasvalem.com}{\texttt{gcnllm.lucasvalem.com}}.
\end{abstract}

\section{Introduction}

The volume of image data being produced and stored has grown dramatically in recent years, driven by applications in areas such as healthcare, security, and social media~\cite{Uelwer2025,Ghosh2024Frontier}. Organizing, indexing, and classifying these large image collections is a central problem for modern data management systems. While collecting images has become inexpensive, labeling them remains costly and time-consuming, as it typically depends on specialized human effort~\cite{Uelwer2025}. This limitation motivates methods that learn effectively from collections in which only a small fraction of the samples is labeled.

In this scenario, semi-supervised methods, which learn jointly from labeled and unlabeled data, have emerged as a promising direction. Among them, Graph Convolutional Networks (GCNs)~\cite{paperGCN-SGC2019} stand out by representing samples as nodes of a graph and propagating label information through the connections between them, allowing the few available labels to guide the classification of the unlabeled ones~\cite{muller2023survey_graph_construction}.

A key obstacle when applying GCNs to image classification is that, unlike citation networks and other naturally graph-structured domains, image collections usually do not come with a predefined graph: it must be constructed from the data itself. Most approaches build this graph from the similarity between feature vectors extracted by pretrained deep networks, typically using kNN or reciprocal kNN graphs~\cite{muller2023survey_graph_construction}. However, how to model an effective graph remains an open research challenge, since proximity in the visual feature space does not always reflect semantic relationships and may introduce noisy connections that degrade classification.

Recently, Large Language Models (LLMs) have shown a remarkable ability to capture high-level semantics from text~\cite{Li2024GraphLLM,Ghosh2024Frontier}, yet their use to support graph construction for image classification is still underexplored~\cite{Li2024GraphLLM}. The objective of this work is to investigate this combination: a Vision Language Model (VLM) generates textual descriptions of the images, and an LLM estimates the semantic similarity between connected images, producing scores that prune semantically irrelevant edges from kNN and reciprocal kNN graphs.
We present this approach together with an experimental evaluation on the Corel5k dataset, whose encouraging preliminary results indicate that LLM-based graph refinement can improve classification accuracy and point to a promising direction for ongoing research.

\section{Proposed Approach}

\begin{figure}[ht!]
    \centering
    \includegraphics[width=\textwidth]{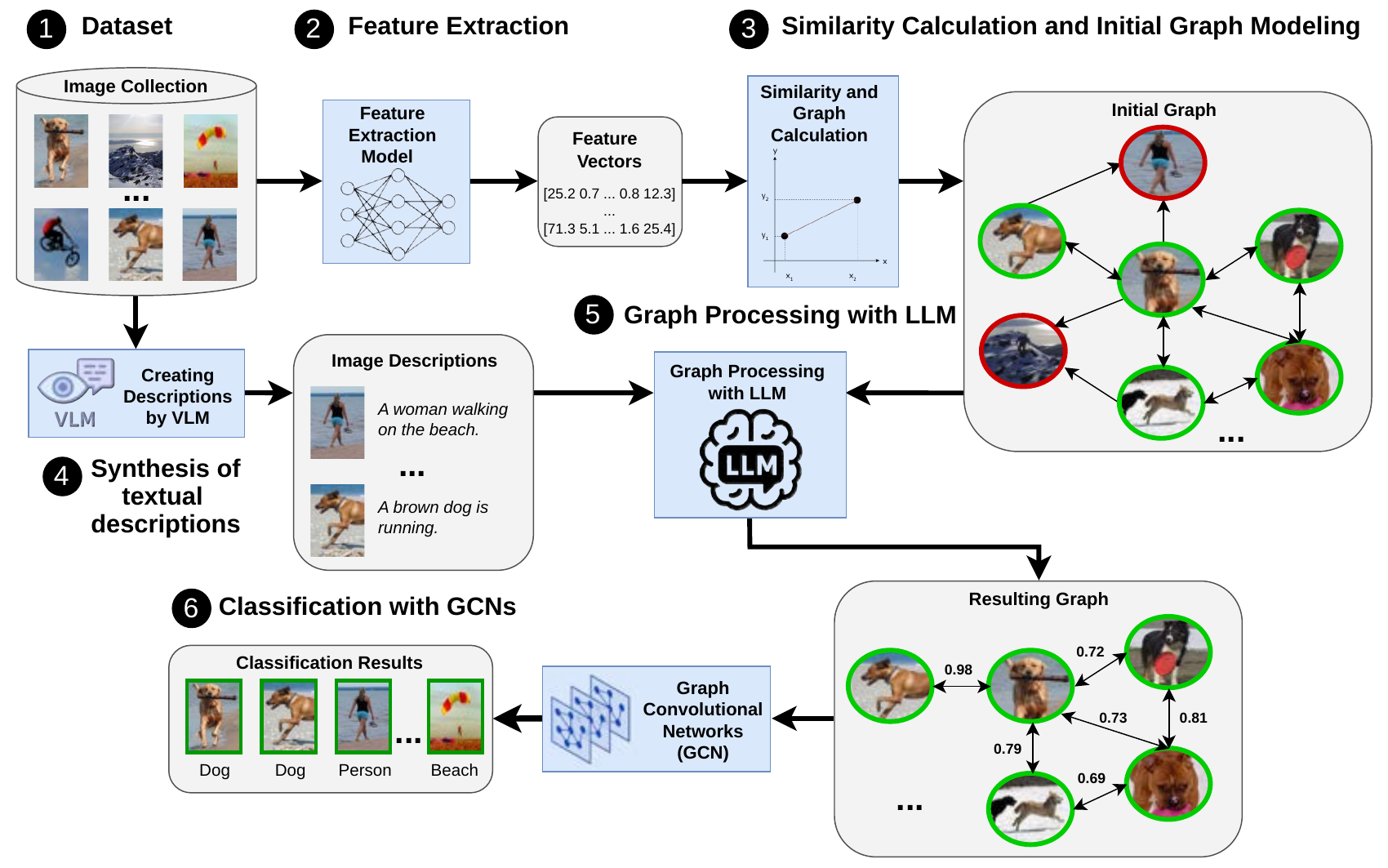}
    \caption{Overview of the proposed approach integrating LLMs for graph refinement with GCNs for semi-supervised image classification.}
    \label{fig:workflow_recuperacao}
\end{figure}

Figure~\ref{fig:workflow_recuperacao} presents an overview of the proposed method for semi-supervised image classification with GCNs, where an LLM refines the graph based on textual descriptions generated from the images by a VLM. The main stages of the process are described in the following subsections.

\subsection{Dataset and Feature Extraction} 
\label{sec:datasets}
The proposed approach is presented using the Corel5k dataset~\cite{PaperCorel5k_PR2013}, which consists of 5,000 images distributed across 50 classes. Since this dataset does not provide associated textual descriptions, the captions required by the method are generated using a VLM  according to the process described in Section~\ref{sec:synthesis}.

For visual feature extraction, the approach employs three pretrained deep learning models: ResNet~\cite{paperRESNET}, the Vision Transformer (ViT)~\cite{dosovitskiy2021vit}, and DINOv2~\cite{oquab2023dinov2}. Each model encodes every image into its own feature vector, and the resulting vectors are used to build a distinct input graph per extractor for the semi-supervised classification with the GCN.

\subsection{Graph Modeling}
\label{sec:graph_model}

Let $x_{i}$ and $x_{j}$ be two images that may or may not be connected. From the feature vectors extracted as described in the previous section, we use a Ball Tree to efficiently retrieve, based on the Euclidean distance, the set of the top-$k$ nearest neighbors of $x_i$, denoted by $\mathcal{N}(x_i, k)$. We consider two common graph construction strategies, which serve both as baselines and as a basis for investigating the use of LLMs: the $k$NN graph and the reciprocal $k$NN graph~\cite{paperManifoldGCN,reID25}.

In the $k$NN graph, each image is connected to its $k$ nearest neighbors, producing the edge set $E_{knn} = \{(i,j) \mid j \in \mathcal{N}(x_i, k)\}$. In the reciprocal $k$NN graph, an edge is created only when the neighborhood relationship is mutual, that is, $E_{rec} = \{(i,j) \mid j \in \mathcal{N}(x_i,k) \,\land\, i \in \mathcal{N}(x_j,k)\}$. The resulting graph can then be refined by the LLM before being provided as input to the GCN.

\subsection{Synthesis of Textual Descriptions} 
\label{sec:synthesis}

Since the Corel5k dataset does not provide textual descriptions associated with the images, a VLM was adopted to generate the captions. In this work, the BLIP (\textit{Bootstrapping Language-Image Pre-training}) model~\cite{Li2022BLIPBL} was employed due to its ability to produce textual descriptions from image content.
The process consists of providing an image as input to the model, which generates a corresponding caption, as illustrated in step 
4 of Figure~\ref{fig:workflow_recuperacao}.
These captions are then employed as textual representations to formulate the input for the LLM during the graph refinement step.

\subsection{Graph Processing with LLM}
\label{sec:llm}

After computing the initial graph as described in Section~\ref{sec:graph_model}, a semantic refinement step is performed using an LLM. For this stage, we adopt the GPT-OSS-20B model, accessed through the Groq platform\footnote{Online API that allows the execution of open-source LLMs: \url{https://groq.com/}.}. This model was chosen because it is a publicly available, open-weight LLM of moderate size, which makes our approach reproducible and computationally accessible without relying on large-scale proprietary models. For reproducibility, all queries used a random seed of $1000$, temperature of $0$, and a maximum generation length of $10{,}000$ tokens.

In this stage, the textual descriptions associated with the images are used to estimate the semantic similarity between connected image pairs, as illustrated in step 5 of Figure~\ref{fig:workflow_recuperacao}. Given the top-$k$ elements of each image provided by a feature extractor (i.e., DINOv2, ResNet, and ViT), we formulate a prompt to compute the semantic similarity between images, as presented in Listing~\ref{prompt}. This prompt instructs the LLM to score how semantically similar each candidate caption is to a reference caption, returning only the numeric scores in the range $[0,1]$. The reference caption corresponds to the query image, while the candidate captions are the captions of its neighbors, sorted in descending order of similarity. This process is repeated for every image to obtain the full set of scores.

The similarity scores generated by the LLM are used to filter connections in the initial graph, retaining only edges whose semantic similarity exceeds a threshold $th$, resulting in the following: $E_{LLM} = \{(i,j) \mid (i,j) \in E_{initial} \land \rho(i,j) \ge th \}$, where $\rho(i,j)$ represents the similarity estimated by the LLM between images $i$ and $j$. Note that $E_{initial}$ can be either $E_{knn}$ or $E_{rec}$.

\begin{lstlisting}[
basicstyle=\footnotesize\ttfamily,
breaklines=true,
frame=single,
caption={Prompt used for semantic similarity evaluation.},
label={prompt}
]
You are evaluating visual semantic similarity.
Reference: {reference_caption}
Step 1: Identify:
* main subject
* key attributes (clothing, objects, actions)
Step 2: Score each phrase:
Scoring rules:
1.0     = same subject + same key attributes
0.8-0.9 = same subject + at least one key attribute
0.5-0.7 = same subject only
0.1-0.4 = weak relation (very generic)
0.0     = different subject or unrelated
Important:
* Focus on visual meaning
* Be consistent across all phrases
Phrases:
1. {neighbor_caption_1}
2. {neighbor_caption_2}
...
k. {neighbor_caption_k}
Output ONLY the scores separated by spaces.
\end{lstlisting}

\subsection{Semi-Supervised Classification with Graph Convolutional Networks}
\label{sec:classification}

The classification stage was performed using the \textit{Simple Graph Convolution} (SGC) model~\cite{paperGCN-SGC2019}. By removing nonlinearities and collapsing the weight matrices into a single linear transformation, SGC is a lightweight and efficient model. This makes it well suited to validate the proposed approach with minimal architectural overhead. The model receives as input the image feature vectors and the graph structure constructed in the previous stage, enabling the integration of visual information and semantic relationships between nodes during the classification process. The idea is that pruning low-similarity edges with an LLM filters out unreliable connections produced by purely visual features, providing a graph that improves classification effectiveness.

\section{Experimental Results}
\label{sec:results}
Two graph construction strategies were evaluated, one based on $k$NN and another on reciprocal $k$NN, both with $k=20$ and including their variants with LLM-based semantic refinement, which is our proposed approach. All configurations were assessed using the SGC model for classification under a reverse stratified 10-fold cross-validation protocol in which only 10\% is used for training and 90\% for testing to assess effectiveness under limited supervision, with classification accuracy adopted as the evaluation measure. All results report the mean and standard deviation of 10 runs.

\begin{figure}[!htb]
    \centering
    \begin{minipage}{\textwidth}
        \centering
        \subfloat[DINOv2 with kNN graph]{\includegraphics[width=0.48\textwidth]{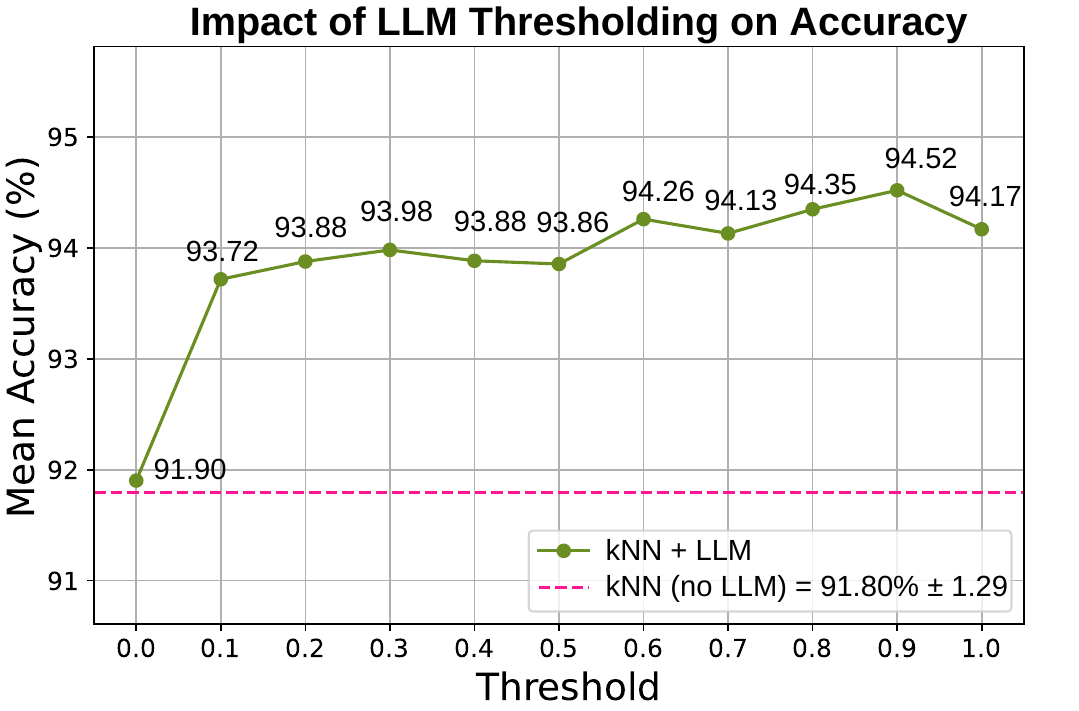}}
        \hfill
        \subfloat[DINOv2 with Rec. kNN graph]{\includegraphics[width=0.48\textwidth]{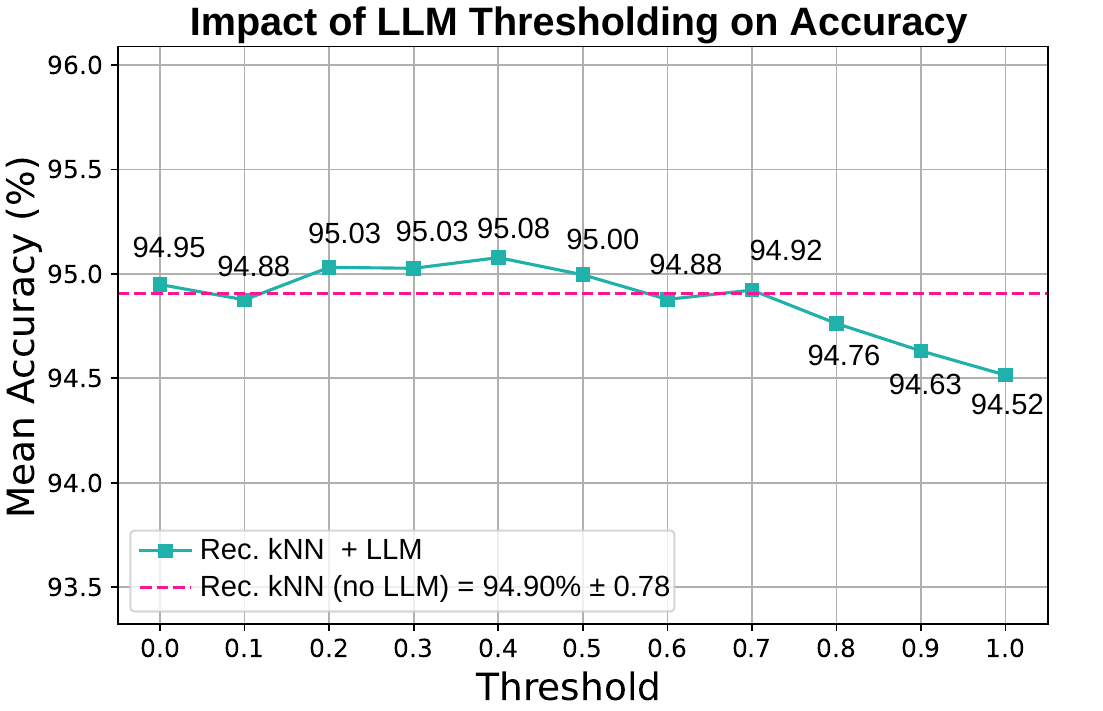}}

        \vspace{0.5em}

        \subfloat[ViT with kNN graph]{\includegraphics[width=0.48\textwidth]{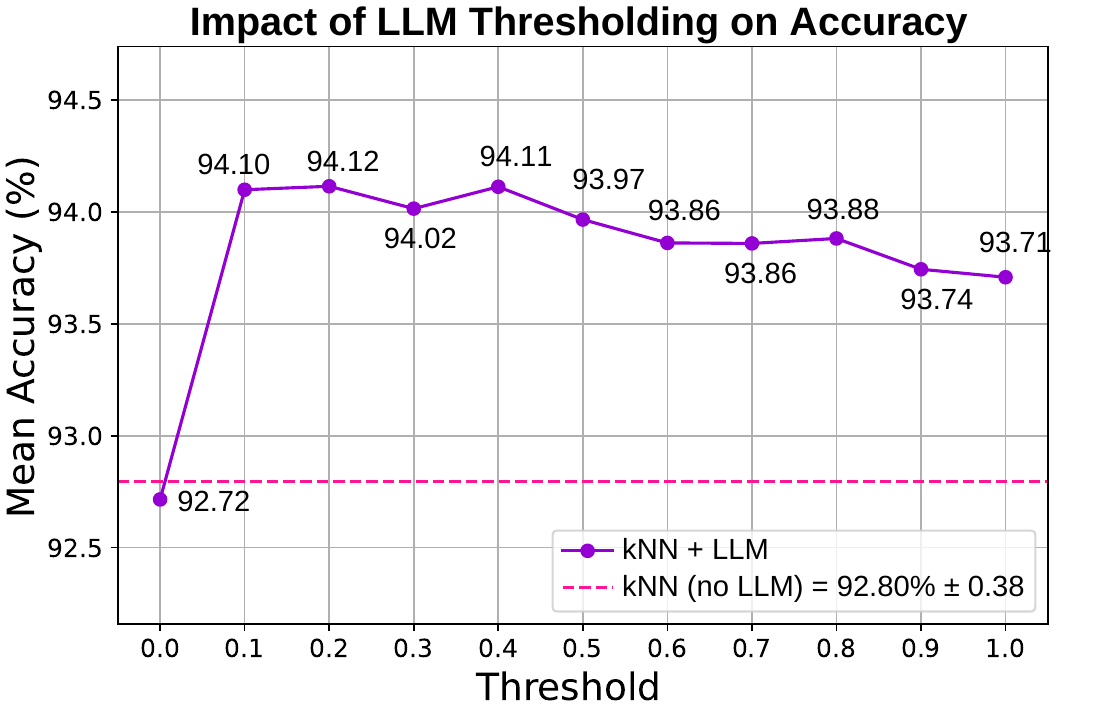}}
        \hfill
        \subfloat[ViT with Rec. kNN graph]{\includegraphics[width=0.48\textwidth]{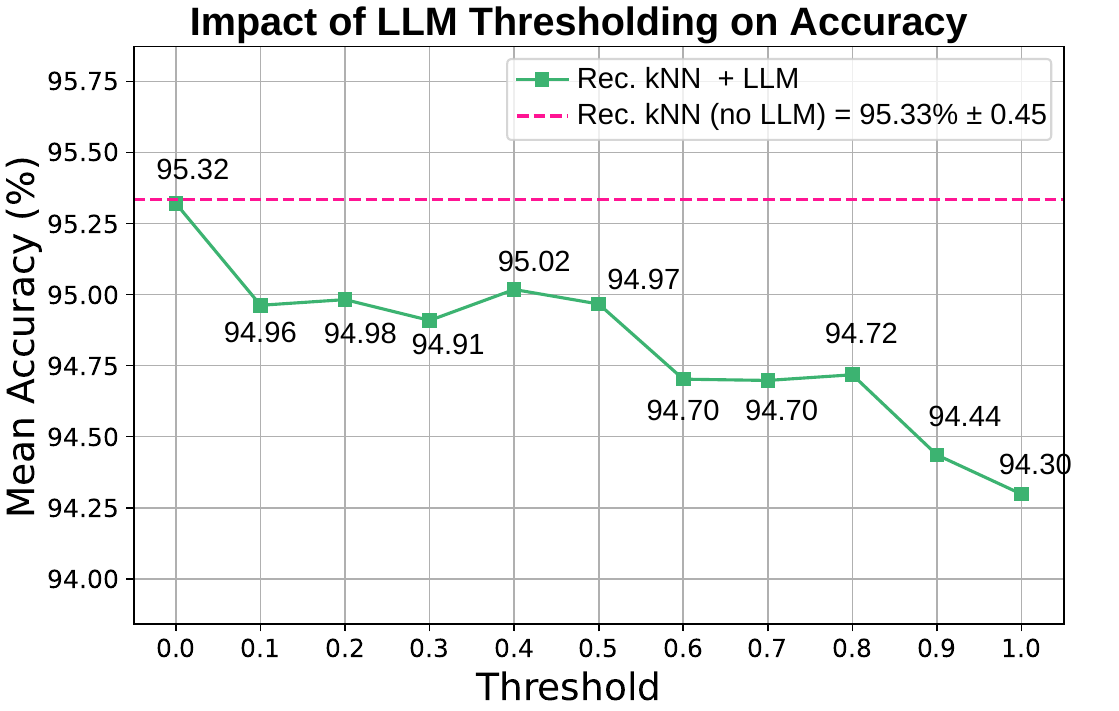}}

        \vspace{0.5em}

        \subfloat[ResNet with kNN graph]{\includegraphics[width=0.48\textwidth]{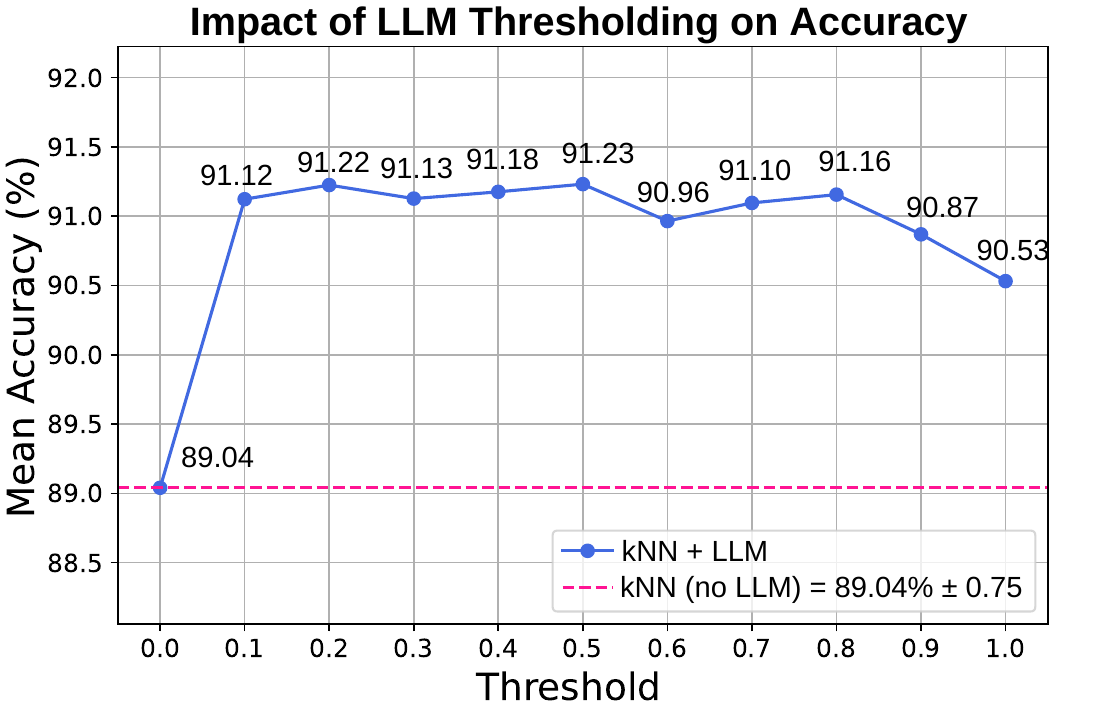}}
        \hfill
        \subfloat[ResNet with Rec. kNN graph]{\includegraphics[width=0.48\textwidth]{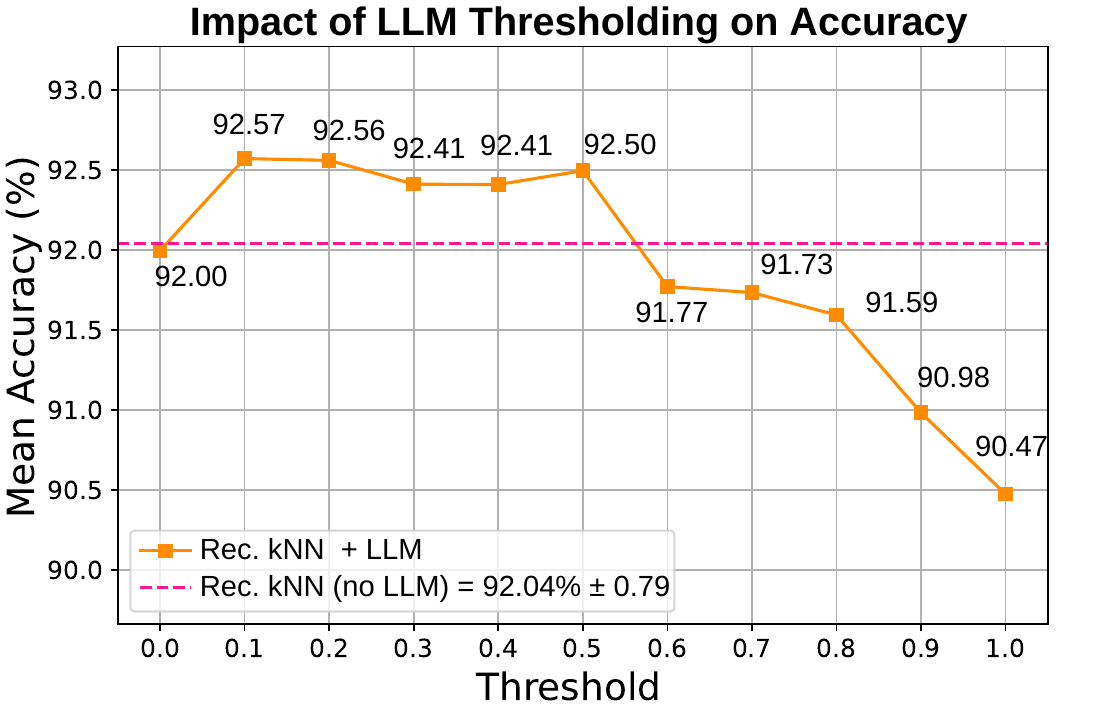}}
    \end{minipage}
    \caption{Impact of LLM thresholding on SGC classification accuracy.}
    \label{fig:graphs}
\end{figure}

To analyze the sensitivity of the refinement step, Figure~\ref{fig:graphs} presents, for each feature and graph combination, a dashed line that indicates the baseline (no LLM) and a curve that shows how the LLM refinement impacts the GCN accuracy as the threshold varies. As can be seen, $0.2$ is a good default value in most cases, and it is therefore adopted in the remaining experiments.

Building on this choice, Table~\ref{tab:graph_results} presents the results obtained for different graphs, reporting both the default threshold of $0.2$ and the best threshold evaluated. Overall, semantic edge refinement based on LLMs improved accuracy in most cases when compared to graphs built solely from visual similarity. The improvements are concentrated on the $k$NN graphs, whose purely visual neighborhoods are often noisier and benefit most from the LLM. In contrast, the reciprocal $k$NN graphs already provide stronger baselines, so the LLM provides smaller gains. A clear saturation case is observed for ViT, which is already a strong feature extractor.

These results suggest that LLM refinement is most effective on noisier graph structures, while providing smaller contributions when the underlying graph is already highly discriminative. Although this behavior deserves further investigation, the consistent gains observed across the noisier configurations highlight the strong potential of LLMs as a flexible tool for enhancing graph-based learning in computer vision tasks, particularly using GCN models. We believe that investigating different prompts, parameters, and LLM models can lead to more improvements, opening promising directions for future research.

\definecolor{DarkGreen}{RGB}{0,125,0}
\definecolor{GrayTone}{RGB}{60,60,60}

\begin{table}[!ht] 
    \centering
    \caption{Classification accuracy (\%) with and without LLM graph refinement.}
    \resizebox{\textwidth}{!}{%
    \begin{tabular}{ | c | l | c | c | c | }
    \hline
    & & \multicolumn{3}{ c | }{\textbf{Feature Extractor}}\\
    \cline{3-5} 
      \multirow[c]{-2}{*}{\textbf{Graph}}  & \multirow[c]{-2}{*}{\textbf{Method}} & \textbf{DINOv2} & \textbf{ViT} & \textbf{ResNet} \\
    \hline
    \multirow{4}{*}{\rotatebox[origin=c]{45}{\textbf{kNN}}}
    & Baseline (no LLM)              & 91.80 $\pm$ 1.29          & 92.80 $\pm$ 0.38          & 89.04 $\pm$ 0.75 \\
    & + LLM (default threshold=0.2)     & 93.88 $\pm$ 1.26          & 94.12 $\pm$ 0.48          & 91.22 $\pm$ 0.60 \\
    & + LLM (with best threshold)  & 94.52 $\pm$ 1.06 (th=0.9) & 94.12 $\pm$ 0.48 (th=0.2) & 91.23 $\pm$ 0.74 (th=0.5) \\
    \cline{2-5}
    & \textit{Relative gain (\%)}    & \textcolor{DarkGreen}{\textbf{+2.96\%}} & \textcolor{DarkGreen}{\textbf{+1.42\%}} & \textcolor{DarkGreen}{\textbf{+2.46\%}} \\
    \hline
    \multirow{4}{*}{\rotatebox[origin=c]{45}{\textbf{Rec. kNN}}}
    & Baseline (no LLM)              & 94.90 $\pm$ 0.78          & 95.33 $\pm$ 0.45          & 92.04 $\pm$ 0.79 \\
    & + LLM (default threshold=0.2)        & 95.03 $\pm$ 0.85          & 94.98 $\pm$ 0.39          & 92.56 $\pm$ 0.55 \\
    & + LLM (with best threshold)   & 95.08 $\pm$ 0.88 (th=0.4) & 95.32 $\pm$ 0.46 (th=0.0) & 92.57 $\pm$ 0.51 (th=0.1) \\
    \cline{2-5}
    & \textit{Relative gain (\%)}    & \textcolor{DarkGreen}{\textbf{+0.19\%}} & \textcolor{GrayTone}{\textbf{$-$0.01\%}} & \textcolor{DarkGreen}{\textbf{+0.58\%}} \\
    \hline
    \end{tabular}%
    }
    \label{tab:graph_results}
\end{table}

\section{Conclusion and Future Work}
\label{sec:conclusion}

This work investigated the use of LLMs to refine graphs for semi-supervised image classification with GCNs. Textual descriptions generated by a VLM are processed by an LLM to estimate the semantic similarity between connected images, pruning edges that are inconsistent in graphs built solely from visual features. Experiments indicate that this refinement can indeed improve accuracy. However, graphs that are already very effective show smaller or no benefit from this strategy, requiring further investigation.
As a continuation of this research, we intend to explore different \textit{prompt engineering}
strategies, additional LLMs, parameter configurations, and other
image datasets to assess our approach, as well as more adaptive ways of combining visual and semantic similarity beyond a fixed threshold.

\section*{Origin of the Work}
This paper presents ongoing work from an undergraduate research project conducted by the student Camila Piscioneri Magalhães.

\section*{Acknowledgments}
Camila Piscioneri Magalhães is supported by a fellowship from the \emph{Programa Unificado de Bolsas} (PUB/USP). This work was also financially supported by the São Paulo Research Foundation (FAPESP, grant \#2025/10602-5), the University of São Paulo (USP, PRPI Ordinance No.~1032, ``\emph{Apoio aos Novos Docentes}''), and the Institute of Mathematics and Computer Science (ICMC-USP).

\section*{AI Usage Declaration}
The authors used AI-based language models (Claude and Gemini) exclusively for reviewing and improving the clarity of text written by the authors.

\bibliographystyle{sbc}
\bibliography{refs}

\end{document}